\documentclass[letterpaper, 10 pt, conference]{IEEEtran}  

\IEEEoverridecommandlockouts                               
\usepackage{cite}
\usepackage{amsmath,amssymb,amsfonts}
\usepackage{algorithmic}
\usepackage{graphicx}
\usepackage{textcomp}
\usepackage{xcolor}
\usepackage{enumerate}
\usepackage[linesnumbered,ruled,lined]{algorithm2e}
\usepackage{booktabs}
\usepackage{threeparttable}
\usepackage{multirow}
\usepackage{stfloats}
\usepackage{subfigure}
\usepackage{graphicx}
\usepackage{rotating}

\usepackage{geometry}
\title{\LARGE \bf
SemanticTopoLoop: Semantic Loop Closure With 3D Topological Graph Based on Quadric-Level Object Map  
}

\author{Zhenzhong Cao$^{1}$, Jingtai Liu$^{1*}$% <-this % stops a space
	\thanks{$^*$The corresponding author of this paper. }
	\thanks{$^{1}$The authors are with the Institute of Robotics and Automatic Information System, College of Artificial Intelligence, Nankai University, Tianjin 300353, China, and also with the Tianjin Key Laboratory of Intelligent Robotics, Nankai University, Tianjin 300350, China (e-mail: 1120230216@mail.nankai.edu.cn; liujt@nankai.edu.cn).}
	\thanks{This work is supported by fundation.}
}

\begin{document}

\maketitle
\thispagestyle{empty}
\pagestyle{empty}

%%%%%%%%%%%%%%%%%%%%%%%%%%%%%%%%%%%%%%%%%%%%%%%%%%%%%%%%%%%%%%%%%%%%%%%%%%%%%%%%
\begin{abstract}
Loop closure, as one of the crucial components in SLAM, plays an essential role in correcting the accumulated errors. Traditional appearance-based methods, such as bag-of-words models, are often limited by local 2D features and the volume of training data, making them less versatile and robust in real-world scenarios, leading to missed detections or false positives detections in loop closure. To address these issues, we first propose a semantic loop closure method based on quadric-level object map topology, which represents scenes through the topological graph of quadric-level objects and achieves accurate loop closure at a wide field of view by comparing differences in the topological graphs. Next, in order to solve the data association problem between frame and map in loop closure, we propose a object-level data association method based on multi-level verification, which can associate 2D semantic features of current frame with 3D objects landmarks of map. Finally, we integrate these two methods into a complete object-aware SLAM system. Qualitative experiments and ablation studies demonstrate the effectiveness and robustness of the proposed object-level data association algorithm. Quantitative experiments show that our semantic loop closure method outperforms existing state-of-the-art methods in terms of precision, recall and localization accuracy metrics.
\end{abstract}

%%%%%%%%%%%%%%%%%%%%%%%%%%%%%%%%%%%%%%%%%%%%%%%%%%%%%%%%%%%%%%%%%%%%%%%%%%%%%%%%
\section{INTRODUCTION}
 Loop closure is indispensable for service robots operating in indoor environments, as they often need to navigate repetitive routes during their long-term operation. Traditional visual SLAM algorithms typically treat loop closure as a scene recognition problem, using low-level 2D features (such as SIFT\cite{lowe2004distinctive}, ORB\cite{rublee2011orb}, etc.) extracted from images for scene recognition and matching, such as the ORB-SLAM series algorithms \cite{mur2015orb, mur2017orb, campos2021orb}, and Vins-Mono \cite{qin2018vins}, based on the DBow2 model \cite{galvez2012bags}. With the advancement of deep learning (e.g., YOLO\cite{redmon2016you}, Mask RCNN\cite{he2017mask}), extracting semantic information from images has become more convenient, leading to the emergence of numerous loop closure methods based on semantic information, which not only can provide richer information for scene recognition, utilizing higher-level scene details, but also can exhibit greater robustness during scene matching, accommodating significant changes in the scene's perspective. 
 % todo: need modify

Due to the abundance of object information in indoor environments, object construction has become mainstream in SLAM processes, leading to the emergence of several excellent object-level SLAM algorithms. Consequently, object-level semantic data association methods and semantic loop closure methods based on object information have also emerged.

However, current object-level semantic data association suffers from issues of low robustness and accuracy in complex scenarios such as false positives and false negatives of object detection network, occlusions and other abnormal conditions. Existing semantic loop closure methods based on object information are prone to false loop detections in repetitive scenes and missed loop detections in scenes with significant perspective changes.

In this paper, to address these challenges, we first introduce a multi-level verification based object-level data association method (abbreviated as \textbf{MLV-ODA} method), which resolves the challenging problem of data association between detetctions of current frame and quadric-level object landmarks of map, particularly in complex scenes where issues like false positives, false negatives, and occlusions are prevalent. This method achieves efficient and accurate data association results and provides prior information for scene representation and matching. Then, we present a quadric-level object map topology based semantic loop closure method (abbreviated as \textbf{QLT-SLC} method), addressing the issues of detecting false positive loop closures under significant viewpoint changes and improving the robustness of loop closure detection. This method enhances both the accuracy and recall of loop closure detection while providing precise loop closure candidates for subsequent global pose correction. Next, we embed the proposed MLV-ODA method and QLT-SLC method into the Object-Aware SLAM system that we proposed in \cite{cao2022object}, forming a complete semantic SLAM system that jointly maintains pose-point-object map database (abbreviated as \textbf{PPO-MD}). Finally, We conduct reasonable qualitative experiments, quantitative experiments, and ablation studies to validate the effectiveness and robustness of the proposed MLV-ODA and QLT-SLC methods in a variety of complex indoor scenes.

\textbf{The main contributions of this work are as follows:}
\begin{itemize}
	\item MLV-ODA method is introduced to reduce the time and space complexity of data association, indirectly promoting the accuracy and completeness of object construction in the scene.
	\item QLT-SLC method is presented to improve the precision and recall rate of loop closure, as well as enhance the system's localization accuracy.
	\item The proposed MLV-ODA method and QLT-SLC method are embed into the Object-Aware SLAM system, which jointly maintain the PPO-MD.
	\item Qualitative experiments, quantitative experiments, and ablation studies are designed to demonstrate the effectiveness and robustness of the proposed MLV-ODA and QLT-SLC method.
\end{itemize}

\begin{figure*}[ht]
    % \vspace{-6mm}
	\centering
	\includegraphics[scale=0.34]{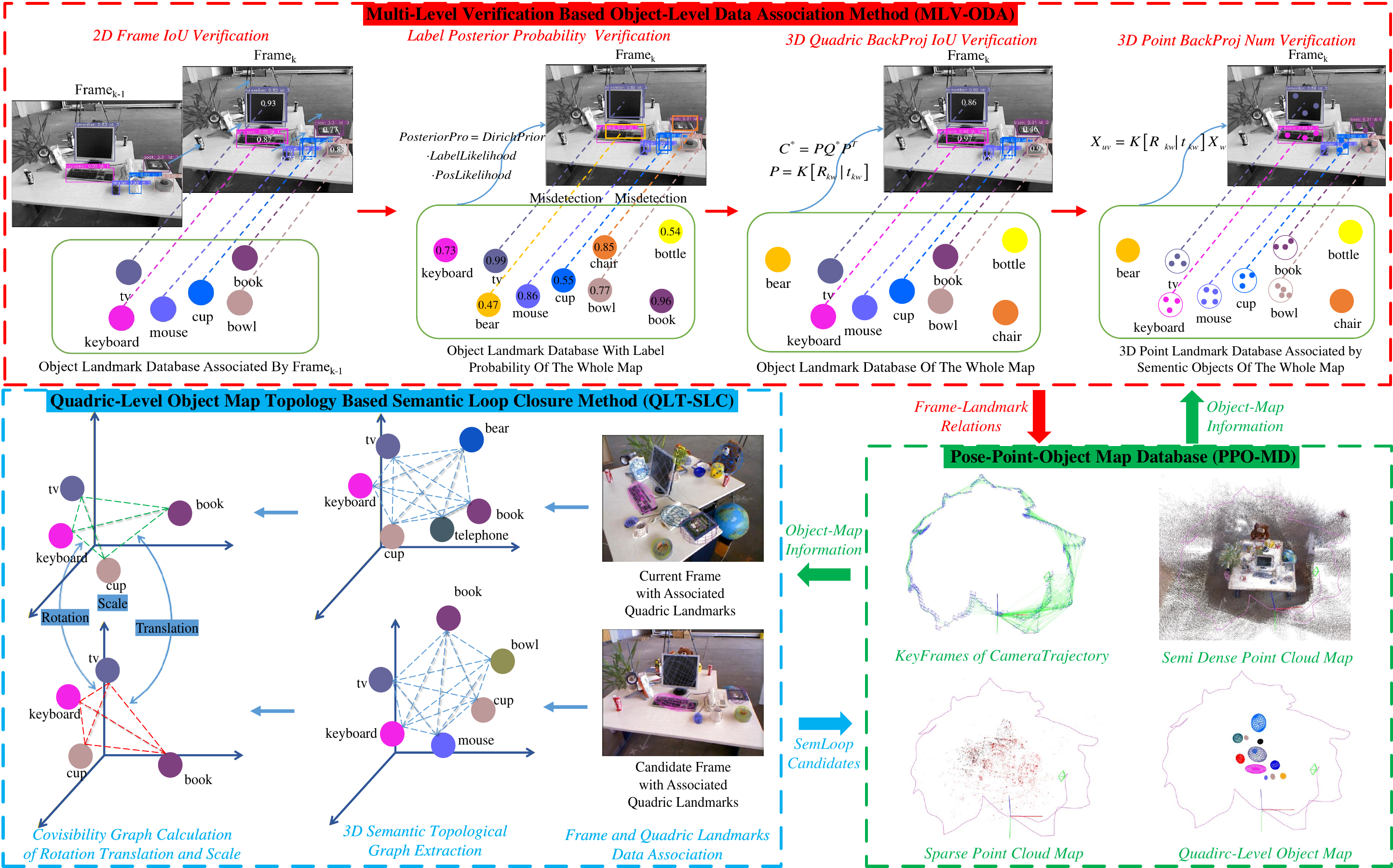}
	\vspace{-2mm}
	\caption{Overview of the proposed system which mainly contains three parts: MLV-ODA method, QLT-SLC method and PPO-MD.}
	\label{overview}
	\vspace{-5mm}
\end{figure*}

\section{Related Work}

\subsection{Semantic Data Association}
% Traditional data association of visual SLAM algorithms refers to associating feature points on images with landmarks in map, such as the ORB feature matching in ORB-SLAM2 \cite{mur2017orb}. With the development of deep learning, some semantic SLAM algorithms have emerged, raising the issue of associating semantic features on images with semantic landmarks in the map.

With the need for object construction in semantic SLAM, some object-level semantic SLAM algorithms have emerged, such as CubeSLAM \cite{yang2019cubeslam} using cubes as object representations and QuadricSLAM \cite{nicholson2018quadricslam} using quadrics. However, these methods did not address the crucial data association issue. Therefore, resolving the data association problem in object-level SLAM became a concern. Wu Y et al. \cite{wu2020eao} built on these two solutions and proposed the ensemble data association algorithm in EAO-SLAM, utilizing non-parametric tests, one-sample t-tests, and two-sample t-tests to indirectly associate objects through point cloud association. However, this algorithm does not leverage the pose and scale information of semantic objects, making it ineffective under conditions of missing or poor-quality point cloud information.

In this context, Tian R et al. \cite{tian2021accurate}, focusing on outdoor scenes, proposed an object data association algorithm that models data association as an assignment problem, constructing multiple association distances containing point cloud and object information and then obtaining the distance matrix by weighted summation, ultimately solving the assignment problem using the Hungarian algorithm \cite{kuhn1955hungarian}. Our previous work \cite{cao2022object} continued this line of thought, improving it to adapt to indoor scenes, and proposed a joint data association algorithm. Although these two solutions partially addressed issues such as occlusion, lighting, and missed detections in complex scenes, as the problem scale increases, the space complexity of data association search and the time complexity of Hungarian algorithm solutions also increase, leading to reduced algorithm efficiency and performance degradation.

\subsection{Semantic Loop Closure}

% Traditional loop closure is typically appearance-based, such as the ORB-SLAM series algorithms \cite{mur2015orb, mur2017orb, campos2021orb}, and Vins-Mono \cite{qin2018vins}, based on the DBow2 model \cite{galvez2012bags}. These methods reconstruct the feature space in the form of an offline dictionary, representing the image scene as a vector of words, compressing the search space for features and speeding up feature matching, thus becoming a common method in the loop closure module of many SLAM systems. However, this bag-of-words-based loop closure has clear shortcomings: on the one hand, it requires a large amount of data for pre-training and has weak scene generalization; on the other hand, relying solely on 2D information for loop closure recognition makes it prone to false positives in similar scenes.

Recently, many solutions propose constructing object-level semantic maps for scenes and then using the parameters of objects in the map and the layout between objects for scene representation, comparing differences between objects for loop detection and loop correction. Lin S et al. \cite{lin2021topology} proposed a loop closure method based on object construction and semantic graph matching, using voxels and cuboids to model object-level features in the environment and further representing the environment as a semantic graph with topological information. Based on this, an efficient graph matching method based on edit distance was proposed for robust location recognition. Finally, loop closure correction was performed through object alignment between semantic graphs. This method can effectively deal with significant viewpoint changes, but when the scene contains duplicate objects with similar topology, graph matching often leads to false alarms. Addressing this issue, Qian Z et al. \cite{qian2022towards} also proposed an approach called SmSLAM+LCD, integrated into the semantic SLAM system, combining high-level 3D semantic information and low-level feature information for accurate loop closure and effective drift reduction. Additionally, Yu J et al. \cite{yu2022semanticloop} proposed SemanticLoop, modeling objects as TSDF and representing the environment as a 3D graph with semantics and topology, which corrects accumulated errors through aligned matching objects. 

Since these methods have achieved very high precision and recall rates in the experiments, hence, we continue the basic ideas of these approaches, aiming to further improve the robustness of loop closure algorithms in complex scenes and their ability to handle significant viewpoint changes, as well as enhance the precision and recall of loop closure method.
% \vspace{-3mm}
\section{System Overview}
Fig.\ref{overview} illustrates the proposed framework, where the MLV-ODA method highlighted in the red box and the QLT-SLC method highlighted in the blue box, which represent the core algorithms of this paper. The PPO-MD in the green box is provided by our previous algorithm \cite{cao2022object}. We have integrated the MLV-ODA method and the QLT-SLC method as two modules into the previous algorithm \cite{cao2022object}, creating a complete object-level semantic SLAM system. However, as the system has been extensively described in the previous paper \cite{cao2022object}, the other modules in the overall framework are omitted. This paper focuses on introducing the MLV-ODA method, the QLT-SLC method, and the PPO-MD.

\textbf{MLV-ODA method:} It involves 2D Frame IoU Verification, Label Posterior Probability Verification, 3D Quadric BackProj IoU Verification, and 3D Point BackProj Num Verification. The four hierarchical verification levels progressively narrow down the search space for data association, aiming to find a landmark association for each detection result as much as possible. At the same time, the association results are also provided to PPO-MD.

\textbf{QLT-SLC method:} It employs Frame and Quadric Landmarks Data Association, 3D Semantic Topological Graph Extraction and Covisibility Graph Calculation of Rotation Translation and Scale to obtain the loop closure candidates, which will be provided to PPO-MD for loop correction.

\textbf{PPO-MD:} It maintains KeyFrames of Camera Trajectory, Semi Dense Point Cloud Map, Sparse Point Cloud Map, and Quadric-Level Object Map. Simultaneously, it provides the required object and point cloud data for the MLV-ODA method and the QLT-SLC method and continuously updates its own database information based on the processing results of the MLV-ODA method and the QLT-SLC method.

\section{The MLV-ODA Method}\label{mlv}

We will introduce the specific roles of each level in our proposed MLV-ODA method. There is a progressive relationship between levels, and each level will filter out the data that can be processed by that level. By adjusting the strategy and search space, we try our best to ensure that each detection can find the corresponding associated landmark.

\subsection{2D-Level Verification For Micromotion Between Frames}
In general, during the operation of SLAM, the system's frame rate is relatively high, and the scene change between adjacent frames is relatively small in the time sequence. Consequently, the movement amplitude of objects in the images is also relatively small. If the same object can be detected in both frames, the bounding box of the object in the two-dimensional pixel level of the image would also have a small movement range, resulting in a relatively large $IoU$. Therefore, we propose 2D Frame IoU Verification to preliminarily associate the current frame with the same object in the previous frame. Since data association has already been completed in the previous frame image, it is possible to indirectly associate the current frame detection results with the quadric landmarks associated with the previous frame. We define $D_k^j$ as the $j$-th detection box in the current frame, $D_{k - 1}^{{Q_i}}$ as the detection box associated with the quadric landmark ${Q_i}$ in the previous frame, and $Io{U_{ij}}$ is defined as follows:
\vspace{-3mm}
\begin{equation}
Io{U_{ij}} = \frac{{D_{k - 1}^{{Q_i}} \cap D_k^j}}{{D_{k - 1}^{{Q_i}} \cup D_k^j}}
\end{equation}
\vspace{-4mm}

For $D_k^j$, we search for its maximum $IoU$ value as follows:

\vspace{-4mm}
\begin{equation}
max(Io{U_{ij}}) = max\{ Io{U_{0j}}, \cdots ,Io{U_{ij}}, \cdots ,Io{U_{nj}}\}
\end{equation}
\vspace{-5mm}

Additionally, for robustness considerations, we implement category validation and threshold validation to further determine the success of data association. Association is considered successful only if condition $Io{U_{ij}} > {\delta _{\rm{1}} }{\rm{ }}$ and ${\rm{ }}Label{\rm{(}}D_k^j{\rm{)}} = Label{\rm{(}}{Q_i}{\rm{)}}$ is satisfied.

\subsection{Pro-Level Verification For False Positive Detection}
Due to the presence of noise in the training dataset, the dataset's incompleteness, and the complexity of the testing environment, the object detection network is prone to false detections. Such false detections can render the 2D Frame IoU Verification ineffective, resulting in failed associations for some objects and subsequently leading to data association interruptions. To address the challenge, we propose the Label Posterior Probability Verification, as referenced in \cite{mu2016slam}. This method involves the probabilistic modeling of the quadric landmarks. Due to the discreteness and uncertainty of the categories, we utilize the Dirichlet process to model the categories of quadric landmarks. According to the bayesian probability model, the posterior probability that the detection box $D_k^j$ belongs to the quadric landmark $Q_i$ can be calculated as follows:

\vspace{-3mm}
\begin{equation}
\begin{array}{c}
PosteriorPr{o_{ij}} \propto DirichletPrior({Q_i}) \cdot \\
       LableLikelihood(D_k^j) \cdot \\
      PosLikelihood(D_k^j,{Q_i})
\end{array}
\end{equation}
\vspace{-3mm}

Through traversal, we find the maximum probability $max(PosteriorPr{o_{ij}})$ that satisfies the condition. If this maximum probability is less than the Dirichlet prior probability that it belongs to a new object, then we give more credibility to this prior probability, indicating that the current detection box should be associated with a new object, resulting in association failure. If this probability is greater than the prior probability of it belonging to a new object, we give more credence to this probability, indicating that the current detection has successfully associated with the object.

\subsection{3D-Level Verification For Leak Detection and Continuity}
Due to the limited scope of the previous two verification methods, which are based only on the detection box data of the previous frame and its associated quadric landmarks, they cannot effectively handle occlusions and lighting conditions that may cause missed detections in the object detection network.

Therefore, to ensure uninterrupted data association and stable long-term system operation, and also considering the issue of search time complexity, we employ a sliding window approach to appropriately expand the search space for quadric landmarks in the data association. Specifically, with the current frame as the reference, we select the nearest M keyframes in the time series and consider the collection of all quadric landmarks associated with them as the search space for the current data association process.

Firstly, for landmarks with existing quadric parameters, we propose the 3D Quadric BackProj IoU Verification. This method involves the back-projection of the quadric parameters of the landmark onto the image to form a projected bounding box. This bounding box is then used to calculate the $IoU$ with the detection bounding box on the image. Let $D_k^j$ denote the $j$-th detection box in the current frame, $ProjD_k^{{Q_i}}$ represent the quadric landmark ${Q_i}$ projected onto the detection box in the current frame. Consequently, $Io{U_{ij}}$ is defined as:

\vspace{-3mm}
\begin{equation}
    Io{U_{ij}} = \frac{{ProjD_k^{{Q_i}} \cap D_k^j}}{{ProjD_k^{{Q_i}} \cup D_k^j}}
\end{equation}
\vspace{-3mm}

The computation of the projected detection box $D_k^{{Q_i}}$ is as follows:

\vspace{-5mm}
\begin{equation}
    D_k^{{Q_i}}{\rm{ = }}BBox(P{Q_i}{P^T})
\end{equation}
\vspace{-5mm}

where $P{\rm{ = }}K[R|t] \in {\Re ^{3 \times 4}}$ represents the projection matrix containing intrinsic and extrinsic parameters, ${Q_i} \in {\Re ^{4 \times 4}}$ is a symmetric matrix with 9 degrees of freedom, $P{Q_i}{P^T}$ represents the projection of the quadric into a conic, and $BBox( \cdot )$ represents the bounding box fitting operator. 

Similar to the approach in section E, we find the maximum value $max(Io{U_{ij}})$ and then use the threshold ${\delta _2}$ to further validate the data association.

Next, for landmarks without quadric parameters, we propose the 3D Point BackProj Num Verification. This involves projecting the associated map points of the landmarks back onto the image, generating projected feature points within different detection boxes, and then calculating the ratio of projected feature points falling into each detection box. We define $D_k^j$ as the $j$-th detection box in the current frame, $Point({Q_i})$ as all projected feature points of the quadric landmark ${Q_i}$, $Point(D_k^j)$ as the projected feature points falling into detection box $D_k^j$. Thus, the projected ratio $Proportio{n_{ij}}$ is defined as:

\vspace{-3mm}
\begin{equation}
    Proportio{n_{ij}}{\rm{ = }}\frac{{Point(D_k^j)}}{{Point({Q_i})}}
\end{equation}
\vspace{-3mm}

Upon finding the maximum value $max(Proportio{n_{ij}})$, we use the threshold ${\delta _3}$ to further validate the data association.

\section{The QLT-SLC METHOD}\label{qlt}

Algorithm \ref{ag1} presents the core data flow of our proposed QLT-SLC method. The algorithm takes the current keyframe and the map database as input and outputs the paired loop closure candidates with the current frame and the best matching score. The processing steps are as follows: First, based on the number of co-observable objects and their ID differences, the algorithm performs an initial filtering of the candidate frames from the map database that meet the criteria, resulting in a set of loop closure candidates. Then, it initializes the optimal loop closure matching pair and the similarity threshold. Next, it traverses the set of loop closure candidates, extracting semantic topological graphs, including semantic nodes and semantic vectors, for each candidate frame and the current frame. Subsequently, it calculates the semantic similarity between the semantic topological graphs of the two frames based on their positions, rotations, and scales. This process is iterated until the best similarity loop closure matching pair is found. Finally, the algorithm evaluates the similarity score threshold and checks for keyframe consistency. If a successful match is determined, the algorithm proceeds to perform loop closure correction based on the current keyframe and the loop closure candidate frame.

\begin{algorithm}[t]
	\caption{Core Data Process Of QLT-SLC Method}
	\label{ag1}
	%\SetAlgoNoLine % ????????
	\KwIn{Current KeyFrame ${I_k}$ and Map Database}
	\KwOut{Best Loop Closure $({I_k},{I_c}^{best},bestscore)$}
	
	// filter candidate keyframes based on objs and ids
	
	${\rm{\{ }}{I_{c}}{\rm{\} }} \leftarrow LoopMatchFilter{\rm{(}}{I_k}{\rm{,}}T{h_{objs}},T{h_{ids}}{\rm{)}}$
	
	% // initialize best keyframe pair and similarity score

        $({I_k},{I_c}^{best},bestscore) \leftarrow {\rm O} $
	
	\For{each ${I_{c}}$ of ${\rm{\{ }}{I_{c}}{\rm{\} }}$}{

            // extract semantic nodes and vectors for keyframes
		
		${\{ \vec v\} _k} \leftarrow 3DSemTopoGraphExtract({I_k})$

            ${\{ \vec v\} _{c}} \leftarrow 3DSemTopoGraphExtract({I_{c}})$

            // compute similarity socre for keyframes

            ${\rm{s}}core \leftarrow CovGraphSimilarityCal({\{ \vec v\} _k}, {\{ \vec v\} _{c}})$
		
		// update best similarity score and best candidate
		
		$bestscore \leftarrow bestscore > score ? bestscore:socre$

            % // update best keyframe candidate

            $I_c^{best} \leftarrow {I_c}$
	}
        \If{$bestscore > T{h_{score}}$}
		{
		    
		    % // determine if it is a continuous loop

                $isTrueLoop \leftarrow ConsistencyCheck{\rm{(}}{I_k},{I_c}^{best}{\rm{)}}$
		    
		    \If{$isTrueLoop$}
		    {
		        // perform final loop correction
		    
		        $LoopClosureCorrection({I_k},{I_c}^{best})$
		    
		    }
		}
\end{algorithm}

\subsection{Loop Closure Candidates Match Preprocessing} 

\begin{figure}[ht]
	\centering
	\includegraphics[scale=0.32]{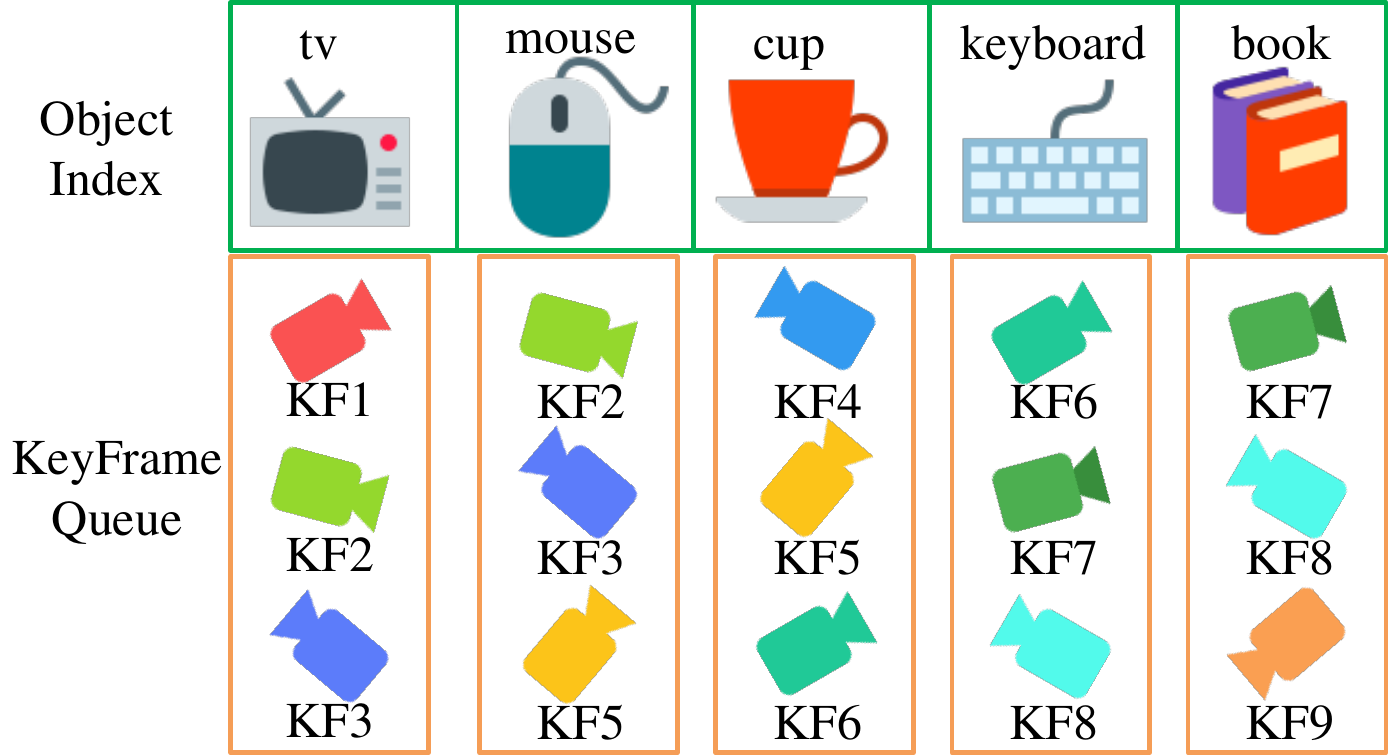}
	\vspace{-3mm}
	\caption{Efficient maintenance strategy based on ObjectIndex and KeyFrameQueue for covisibility information.}
	\label{lm}
	\vspace{-5mm}
\end{figure}

Through MLV-ODA, we have obtained association results between detections and landmarks. Before matching the candidate keyframes for the current keyframe, we first preliminarily filter whether there are enough co-observed landmarks and sufficient ID differences between the two keyframes. However, Directly comparing the differences between two keyframes would result in a high time complexity and relatively low efficiency. Therefore, we propose an efficient maintenance strategy, as shown in Fig.\ref{lm}. The first column represents the ObjectIndex, which maintains landmarks of different categories to quickly index the same object. The second column represents the KeyFrameQueue, which stores the queue of keyframes that observe the corresponding object. Whenever a keyframe is inserted into the map, each data association result in the keyframe is traversed. If the indexed object landmark exists in the ObjectIndex, the corresponding KeyFrameQueue is updated with the current keyframe. Otherwise, a new ObjectIndex and KeyFrameQueue are created for maintenance.

With this maintenance strategy, we can quickly and efficiently find a candidate keyframe set for the current keyframe, where the number of co-observed object landmarks is greater than or equal to $T{h_{objs}}$ and the ID difference between the keyframes is greater than or equal to $T{h_{ids}}$.
		
\subsection{3D Semantic Topological Graph Extraction} 

After obtaining the candidate keyframe set, we proceed to extract the 3D semantic topological graph for each keyframe, which includes 3D semantic nodes and 3D semantic vectors. The specific operations are as follows:

Let ${I_i}$ represent the current keyframe, ${I_j}$ represent the candidate keyframe, and $\{ (Q_p^i,Q_q^j)\}$ represent the set of quadric-level semantic nodes associated with ${I_i}$ and ${I_j}$. Let $C_p^j = \left( {X_p^j,Y_p^j,Z_p^j} \right)$ be the center point coordinates of semantic node $Q_p^i$ in the ${I_i}$ coordinate system, and $C_q^j = \left( {X_q^j,Y_q^j,Z_q^j} \right)$ be the center point coordinates of semantic node $Q_q^i$ in the ${I_j}$ coordinate system. Here, the topological structure between quadric landmarks is defined as the semantic vector formed by the lines connecting the center point coordinates of each pair of semantic nodes. For example, ${\vec v} = \overrightarrow {C_1^kC_2^k}$ represent the semantic vector between the 1st and 2nd semantic nodes in the ${k}$-th frame.

The semantic vector between the 1st and 2nd semantic nodes in the current keyframe ${I_i}$ is:

\vspace{-3mm}
\begin{equation}
    {\vec v_i}  = {({v_{ix}},{v_{iy}},{v_{iz}})^T} =\overrightarrow {C_1^iC_2^i}
\end{equation}
\vspace{-5mm}

The semantic vector between the 1st and 2nd associated semantic nodes in the candidate keyframe ${I_j}$ is:

\vspace{-3mm}
\begin{equation}
{\vec v_j} = {({v_{jx}},{v_{jy}},{v_{jz}})^T} =  \overrightarrow {C_1^jC_2^j}
\end{equation}
\vspace{-5mm}

According to the properties of vectors in space, the semantic vectors ${v_1}$ and ${v_2}$ differ by a rotation ${R}$ and scale ${s}$, that is:

\vspace{-3mm}
\begin{equation}
{\vec v_j} = R(s{\vec v_i})
\end{equation}
\vspace{-5mm}

Here, $R \in {\Re ^{3 \times 3}}$ represents the rotation matrix between the two vectors, and ${s}$ is a scalar representing the scale factor between the two vectors. We can solve for ${R}$ and ${s}$ using the following Rodrigues’s Formula:

\vspace{-3mm}
\begin{equation}
R = \cos \theta I + (1 - \cos \theta )\vec \omega {\vec \omega ^T} + \sin \theta {\vec \omega ^ \wedge }
\end{equation}
\vspace{-5mm}

\vspace{-3mm}
\begin{equation}
s = \frac{{\left| {{{\vec v}_j}} \right|}}{{\left| {{{\vec v}_i}} \right|}}
\end{equation}
\vspace{-3mm}

Where $\theta$ are the rotation angle, and ${{\vec \omega} = ({\omega _x},{\omega _y},{\omega _z})^T}$ is the rotation axis. The calculation of the rotation angles and rotation axis is as follows:

\vspace{-3mm}
\begin{equation}
\theta  = \arccos \frac{{{{\vec v}_i}{{\vec v}_j}}}{{\left| {{{\vec v}_i}} \right|\left| {{{\vec v}_j}} \right|}}
\end{equation}
\vspace{-3mm}

\vspace{-3mm}
\begin{equation}
\vec \omega  = \left( {\begin{array}{*{20}{c}}
{{\omega _x}}\\
{{\omega _y}}\\
{{\omega _z}}
\end{array}} \right) = \left( {\begin{array}{*{20}{c}}
{{v_{iy}}{v_{jz}} - {v_{iz}}{v_{jy}}}\\
{{v_{iz}}{v_{jx}} - {v_{ix}}{v_{jz}}}\\
{{v_{ix}}{v_{jy}} - {v_{iy}}{v_{jx}}}
\end{array}} \right)
\end{equation}
\vspace{-3mm}

In addition to considering the rotation and scale between the semantic vectors, we also consider the gap ${\vec t}$ between the starting points of the semantic vectors. Considering the starting point of the semantic vector ${\vec v_i}$ as $C_1^i$ and the starting point of ${\vec v_j}$ as $C_1^j$, the corresponding translation ${\vec t}$ is:

\vspace{-3mm}
\begin{equation}
{\vec t} = {(X_1^j - X_1^i,Y_1^j - Y_1^i,Z_1^j - Z_1^i)^T}
\end{equation}
\vspace{-5mm}

Thus, we have established the three indicators, ${R}$, ${s}$, and ${t}$, to measure the similarity between two semantic vectors. When the two semantic vectors coincide, the three indicators are $R = {I_{3 \times 3}}$, $s = 1$, and $t = {(0,0,0)^T}$, where ${I}$ is the unit matrix.

\subsection{Covisibility Graph Similarity Calculation} 

Based on the definitions and transformation relationships provided for the semantic vectors, we can proceed with the similarity calculation for the covisibility graph in the 3D semantic topological graph. The specific steps are as follows:

Initialize the spatial semantic similarity ${Score = 0}$.

Traverse the semantic node set of the current keyframe ${I_i}$and the candidate keyframe ${I_j}$.

Calculate the rotation, scale, and translation indicators ${{\{R,s,\vec t\}}}$ between the semantic vectors formed by each pair of semantic nodes in the current keyframe and the corresponding two semantic nodes in the candidate keyframe.

Calculate the topological structure similarity score between the two semantic vectors:

\vspace{-3mm}
\begin{equation}
Scor{e_{topology}} = {{\mathop{\rm e}\nolimits} ^{ - \frac{1}{2}\left| {1 - s\left\| R \right\| - \left\| \vec t \right\|} \right|}}
\end{equation}
\vspace{-5mm}

Update the semantic similarity score:
\vspace{-2mm}
\begin{equation}
Scor{e_{semantic}} = Scor{e_{semantic}} + \rho Scor{e_{topology}}
\end{equation}
\vspace{-5mm}

Update the score of the spatial semantic similarity between the topological graph:
\vspace{-2mm}
\begin{equation}
Score = Score + Scor{e_{semantic}}
\end{equation}
\vspace{-5mm}

After traversal is completed, output the final spatial semantic similarity score between the two topological graphs.

We use formula (11) to calculate the similarity between two semantic vectors in the topological graph. Its characteristic is that as ${s\left\| R \right\| + \left\| {\vec t} \right\|}$ approaches 1, the function value increases faster and faster, so it is easier to process than linear The function is more strict because we hope that the three indicators between two semantic vectors will have a higher similarity score only when they are close to ${\{ \left\| R \right\| = 1,s = 1,\left\| {\vec t} \right\| = 0\}}$.

In addition, this paper presents ${\rho Scor{e_{topology}}}$ when calculating the semantic similarity score using formula (16), where $\rho$ represents the quality score of the quadric landmark corresponding to the starting semantic node of the semantic vector in the topological graph (which is obtained from the EQI algorithm in the previous work \cite{cao2022object}). The purpose of using the value $\rho$ is that when the parameter quality score of the constructed quadric landmark is relatively small, it means that the confidence of the quadric landmark is relatively low, so its impact on loop closure detection should be smaller.

\begin{figure*}[ht]
	\centering
	\includegraphics[scale=0.2]{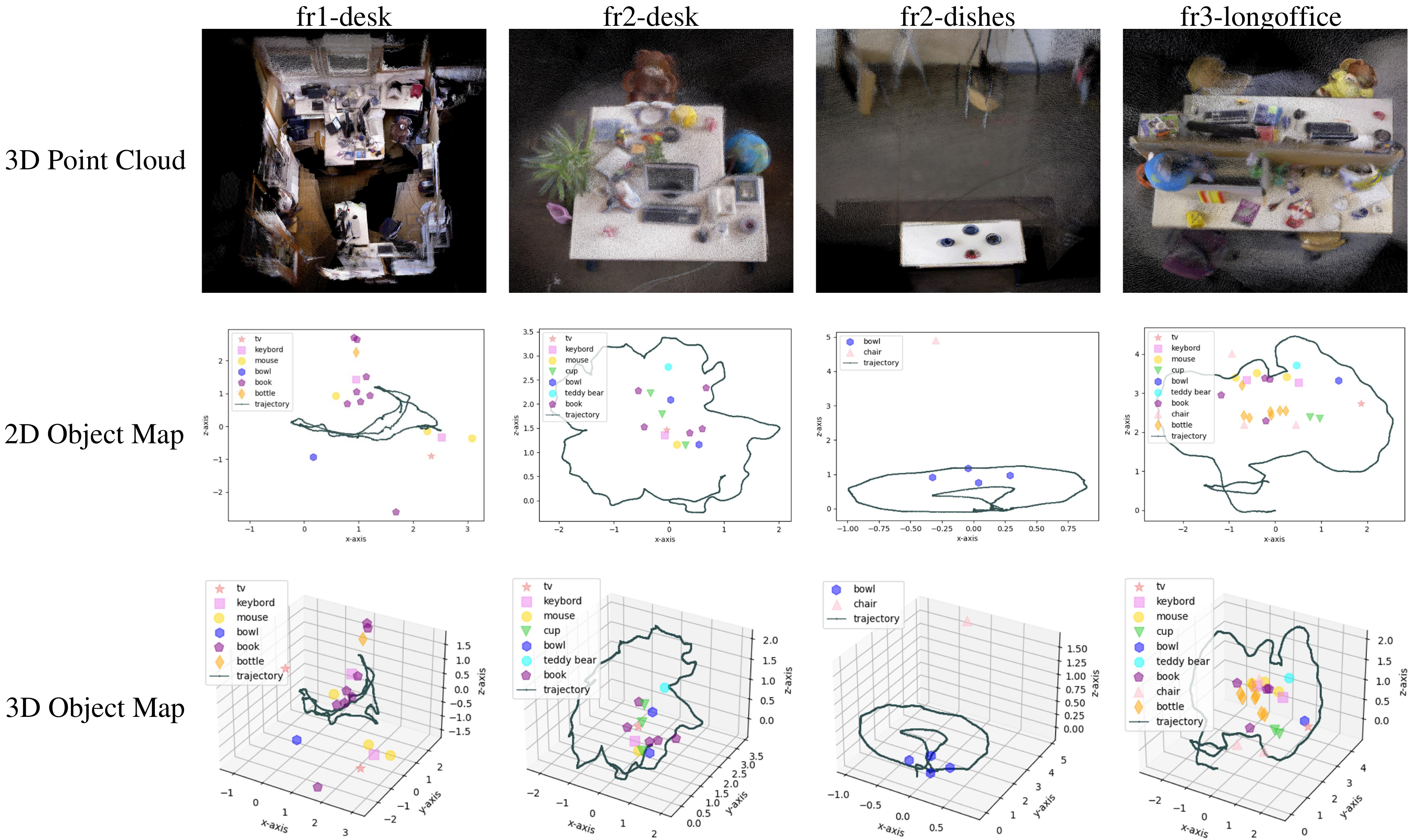}
	\vspace{-3mm}
	\caption{Qualitative performance of semantic objects construction in four different indoor scenes in the TUM dataset.}
	\label{oda1}
	\vspace{-2mm}
\end{figure*}

\begin{figure*}
	\centering
	\includegraphics[scale=0.2]{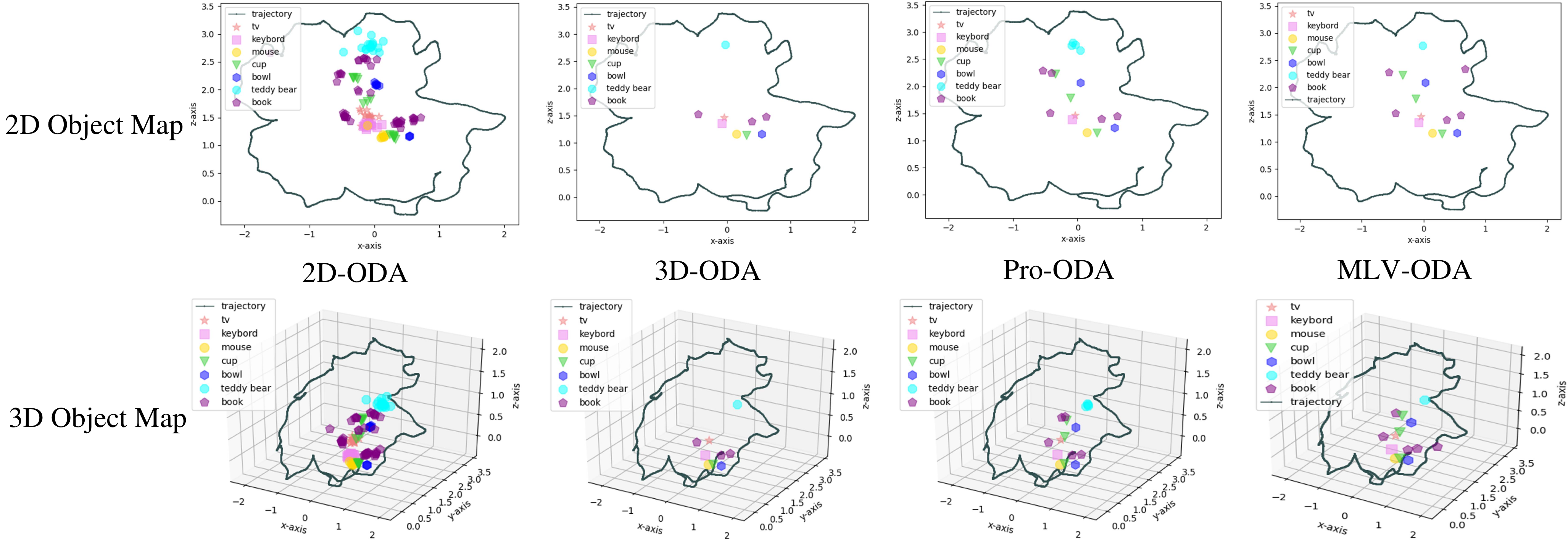}
	\vspace{-3mm}
	\caption{Ablation study performance on semantic object construction in the TUM dataset fr2-desk sequence.}
	\label{oda2}
	\vspace{-5mm}
\end{figure*}

\section{Experiments and Evaluation}
We evaluated our proposed method using the TUM RGB-D benchmark \cite{sturm2012benchmark}, which is a commonly used benchmark for evaluating the performance of visual SLAM algorithms. All experiments were conducted on our local device, with the following specifications: AMD Ryzen 7 5800H with Radeon Graphics 3.20 GHz, 16.0 GB RAM, Nvidia RTX 3060 6GB.

\subsection{The Performance of MLV-ODA Method}

Data association performance cannot be measured directly, but it impacts the semantic objects construction performance. Therefore, we indirectly validate the effectiveness and robustness of the proposed MLV-ODA method using the performance of semantic objects construction. We introduce indicators precision and recall to measure the accuracy and completeness of semantic object construction respectively. We judge that only when the category of the object is consistent with the real scene and there are no other constructed objects in the vicinity of the object, it can be considered a true positive.

Based on the performance shown in both the Table \ref{t1} and the Fig.\ref{oda1}, our proposed method exhibits favorable results. Qualitatively, our method accurately constructs semantic objects in the environment, including poses and labels, with minimal occurrences of erroneous construction, missing construction, or redundant construction. Quantitatively, our method achieves a high level of performance, with an average precision of 97.49$\%$, an average recall of 85.88$\%$, and an average F-value of 91.31$\%$.

\begin{table*}[]
\begin{center}
\caption{Precision and recall experiment of our MLV-ODA method on four sequences of TUM dataset}\label{t1}
\vspace{-2mm}
\begin{tabular}{clccccc}
\hline
\multicolumn{2}{c}{Sequence}                  & fr1-room & fr2-desk & fr2-dishes & fr3-longoffice & Average \\ \hline
\multicolumn{2}{c}{Num of Sequence}           & 1352     & 2893     & 1706       & 2488           & 2109.75 \\
\multicolumn{2}{c}{Num of Constructed Object} & 17       & 14       & 5          & 24             & 15      \\
\multicolumn{2}{c}{Num of RealExist Object}   & 19       & 16       & 6          & 26             & 16.75   \\
\multicolumn{2}{c}{True Positive}             & 16       & 14       & 5          & 23             & 14.5    \\
\multicolumn{2}{c}{Precision}                 & 94.12\%  & 100.00\% & 100.00\%   & 95.83\%        & 97.49\% \\
\multicolumn{2}{c}{Recall}                    & 84.21\%  & 87.50\%  & 83.33\%    & 88.46\%        & 85.88\% \\
\multicolumn{2}{c}{F-Measure}                 & 88.89\%  & 93.33\%  & 90.91\%    & 92.00\%        & 91.31\% \\ \hline
\end{tabular}
\end{center}
\vspace{-5mm}
\end{table*}

\begin{table*}[]
\begin{center}
\caption{Precision and recall ablation study on fr2-desk sequence of TUM dataset}\label{t2}
\vspace{-2mm}
\begin{tabular}{clcccc}
\hline
\multicolumn{2}{c}{Method}                    & 2D-ODA & 3D-ODA   & Pro-ODA & Our MLV-ODA       \\ \hline
\multicolumn{2}{c}{Num of Constructed Object} & 125    & 9        & 17      & \textbf{14}       \\
\multicolumn{2}{c}{Num of RealExist Object}   & 16     & 16       & 16      & \textbf{16}       \\
\multicolumn{2}{c}{True Positive}             & 1      & 9        & 13      & \textbf{14}       \\
\multicolumn{2}{c}{Precision}                 & 0.80\% & 100.00\% & 76.47\% & \textbf{100.00\%} \\
\multicolumn{2}{c}{Recall}                    & 6.25\% & 56.25\%  & 81.25\% & \textbf{87.50\%}  \\
\multicolumn{2}{c}{F-Measure}                 & 1.42\% & 72.00\%  & 78.79\% & \textbf{93.33\%}  \\ \hline
\end{tabular}
\end{center}
\vspace{-7mm}
\end{table*}

To further validate the effectiveness of the proposed hierarchical approach in the MLV-ODA method, we conducted ablation experiments by isolating different levels within the MLV-ODA method. Specifically, we extracted three data association methods: 2D-ODA, which exclusively uses 2D-level verification, 3D-ODA, which solely relies on 3D-level verification, and Pro-ODA, which employs only probabilistic-level verification. We then compared these methods with our MLV-ODA algorithm.

As observed from the performance in the Table \ref{t2} and the Fig.\ref{oda2}, our algorithm achieves the best Precision and Recall, highlighting the effectiveness of the proposed multi-level verification approach. The 2D-ODA method exhibits the poorest performance mainly due to the instability of the object detection network, resulting in a high frequency of interruptions in data association due to numerous false positives and false negatives. Although the 3D-ODA method achieves the same Precision as MLV-ODA, it suffers from lower recall. This is because although 3D-ODA can handle some cases of missed detections and occlusions, insufficient data on object association due to observation limitations leads to failed semantic objects construction. However, the Pro-ODA method performs comparably to our proposed MLV-ODA, primarily due to the integration of semantic category and object position information. Nevertheless, the assumed probabilistic model may not be universally applicable to all objects, resulting in some unsuccessful associations.
% Please add the following required packages to your document preamble:
% \usepackage{multirow}
\begin{table}[]\label{ttime}
\begin{center}
\caption{Comparison of average time consumption between JDA and MLV-ODA at various stages of operation(ms)}\label{t2}
\vspace{-2mm}
\begin{tabular}{cccc}
\hline
Sequenece                       & Metric & JDA    & MLV-ODA \\ \hline
\multirow{3}{*}{fr2-desk}       & 10     & 0.0758 & 0.0454  \\
                                & 100    & 0.1867 & 0.0420  \\
                                & 1000   & 2.8692 & 0.0418  \\ \hline
\multirow{3}{*}{fr3-longoffice} & 10     & 0.0128 & 0.0812  \\
                                & 100    & 0.1907 & 0.0732  \\
                                & 1000   & 2.9550 & 0.0450  \\ \hline
\end{tabular}
\end{center}
\vspace{-7mm}
\end{table}

In addition, we also tested the efficiency of our MLV-ODA method compared with the JDA method using the Hungarian algorithm that we previously proposed in \cite{cao2022object}. In order to make the test intuitive, we recorded the method running time in three stages (10 frames, 100 frames and 1000 frames). The data in Table \ref{ttime} can be seen that the time consumption of our method in each stage is not much different and is very stable. However, due to the time complexity of the Hungarian algorithm, the time consumption of the JDA method will surge as the number of frames increases.

\subsection{The Performance of QLT-SLC Method}
\begin{figure}[ht]
    % \vspace{-6mm}
	\centering
	\includegraphics[scale=0.27]{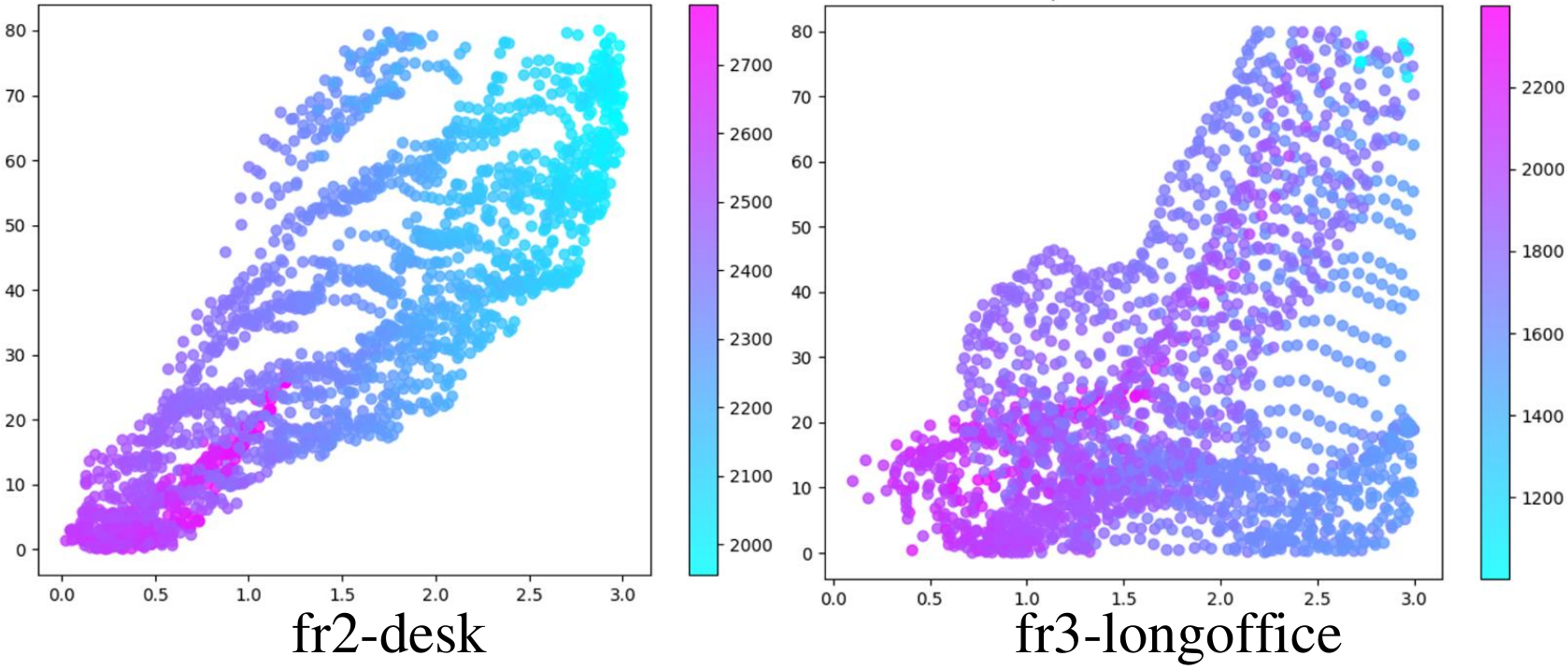}
	\vspace{-3mm}
	\caption{Reference loop closure statistics under condition: position difference within 3m, angle difference within 80° and ID difference above 1000}
	\label{loopclosure}
	\vspace{-1mm}
\end{figure}

We selected two recent outstanding approaches, \cite{qian2022towards} and \cite{yu2022semanticloop}, as the comparative methods for our QLT-SLC method in this paper. However, their criteria for selecting reference loop closures are different. The criteria proposed in \cite{qian2022towards} for selecting reference loop closure candidates are as follows: a position difference of within 1m, an angle difference of within 53°, and an ID difference of over 1000. The criteria proposed in \cite{yu2022semanticloop} are based on the condition that the number of commonly observed objects between two frames is greater than 2, and the ID difference is over 500. However, in our practice, using the criteria proposed in \cite{qian2022towards}, we do not can obtain a similar number of reference loop closures. Therefore, for the fairness, after our experiment, we use the following selection criteria: a position difference of within 3m, an angle difference of within 80°, and an ID difference of over 1000, which actually adds difficulty to our method. The Fig.\ref{loopclosure} shows the statistical graphs of selected reference loops in the fr2-desk and fr3-longoffice sequences of the TUM dataset for our method.

The Table \ref{t3} presents the comparative experiments of our method and three other methods, including traditional loop closure methods \cite{mur2017orb} and \cite{campos2021orb}, as well as semantic loop closure methods \cite{qian2022towards} and \cite{yu2022semanticloop}. Among them, Ours-C1 represents the result of our method under the matching configuration of \cite{qian2022towards}, and Ours-C2 represents the result of our method under the matching configuration of \cite{yu2022semanticloop}. The main comparison metrics are common in loop closure, namely precision and recall. From the table, we observe that our QLT-SLC method, under both configurations, outperforms the original algorithms. Additionally, our method not only detects more loop closures but also exhibits robustness under significant viewpoint changes.

The reason why our method performs well is that on the one hand, thanks to the MLV-ODA method, the data association between the current frame and the map is very robust. We can find more object information for the current frame, so that even under large viewing angle changes, we can find a lot of TP loop closures. On the other hand, the similarity calculation we proposed combines spatial information and semantic information, as well as loop closure consistency check, which can avoid FP loop closures caused by similar image scenes.

\subsection{The Performance of Localization}
 We use the ATE metric to do localization accuracy comparison experiments. Data for \cite{mur2017orb}, \cite{campos2021orb}, and our method were generated on our local device, while the data for the three approaches \cite{lin2021topology}, \cite{qian2022towards} and \cite{yu2022semanticloop} were obtained directly from the respective papers as they were not publicly available.

As shown in the Table \ref{t4}, since our method can detect more accurate loop closure, therefore, our method exhibits the best performance. Even compared with the latest approach \cite{yu2022semanticloop}, achieving competitive accuracy results.

\begin{table*}[]
\begin{center}
\caption{Comparative experiment on precision and recall of loop closure on two sequences of TUM dataset}\label{t3}
\vspace{-2mm}
\begin{tabular}{cccccccc}
\hline
Sequence                       & Metrics       & ORB-SLAM2   & ORB-SLAM3   & \cite{qian2022towards}     & \cite{yu2022semanticloop}  & Ours-C1            & Ours-C2            \\ \hline
\multirow{6}{*}{fr2-desk}      & KeyFrame      & 193         & 261         & /           & /        & \textbf{192}      & /                 \\
                               & ReferenceLoop & 1697        & 2774        & 2122        & /        & \textbf{2079}     & /                 \\
                               & RealLoop      & 176         & 361         & 23          & 44       & \textbf{45}       & \textbf{55}       \\
                               & True Positive & 33          & 66          & 23          & 44       & \textbf{45}       & \textbf{55}       \\
                               & Precision     & 18.75\%     & 18.28\%     & 100.00\%    & 100.00\% & \textbf{100.00\%} & \textbf{100.00\%} \\
                               & Recall        & 1.94\%      & 2.38\%      & 1.08\%      & /        & \textbf{2.16\%}   & /                 \\ \hline
\multirow{6}{*}{fr-longOffice} & KeyFrame      & 239         & 293         & /           & /        & \textbf{274}      & /                 \\
                               & ReferenceLoop & 1773        & 2654        & 2429        & /        & \textbf{2188}     & /                 \\
                               & RealLoop      & 209         & 504         & 6           & 76       & \textbf{25}       & \textbf{85}       \\
                               & True Positive & 16          & 53          & 6           & 76       & \textbf{25}       & \textbf{85}       \\
                               & Precision     & 7.66\%      & 10.52\%     & 100.00\%    & 100.00\% & \textbf{100.00\%} & \textbf{100.00\%} \\
                               & Recall        & 0.90\%      & 2.00\%      & 0.25\%      & /        & \textbf{1.14\%}   & /                 \\ \hline
\multicolumn{2}{c}{Configuration}              & 3m,80°,1000 & 3m,80°,1000 & 1m,53°,1000 & 2,500    & 3m,80°,1000       & 2,500             \\ \hline
\end{tabular}
\end{center}
\vspace{-5mm}
\end{table*}

% Please add the following required packages to your document preamble:
% \usepackage{multirow}
\begin{table*}[]
\begin{center}
\caption{Comparative experiment of localization accuracy after loop closure on TUM dataset(M)}\label{t4}
\vspace{-2mm}
\begin{tabular}{cccccccc}
\hline
Sequence               & Metrics & ORB-SLAM2 & ORB-SLAM3 & \cite{lin2021topology} & \cite{qian2022towards} & \cite{yu2022semanticloop} & Ours            \\ \hline
\multirow{3}{*}{fr2-desk}       & RMSE    & 0.008     & 0.018     & 0.014   & 0.008   & 0.042   & \textbf{0.0078} \\
                                & MEAN    & 0.008     & 0.017     & /       & /       & /       & \textbf{0.0072} \\
                                & MEDIAN  & 0.007     & 0.015     & /       & /       & /       & \textbf{0.0068} \\
\multirow{3}{*}{fr3-longoffice} & RMSE    & 0.012     & 0.012     & 0.016   & 0.009   & 0.015   & \textbf{0.0087} \\
                                & MEAN    & 0.011     & 0.011     & /       & /       & /       & \textbf{0.0079} \\
                                & MEDIAN  & 0.009     & 0.010     & /       & /       & /       & \textbf{0.0075} \\ \hline
\end{tabular}
\end{center}
\vspace{-7mm}
\end{table*}

% \vspace{-1mm}
\section{Conclusion}
In this paper, to address the issue of inaccurate data association under scenarios such as false positives, false negatives, and occlusions, we propose the MLV-ODA method, which reduces the time and space complexity of solving the data association problem, thereby improving efficiency while ensuring accuracy. To tackle the problem of non-robust loop closure detection in scenarios involving significant viewpoint changes or similar scenes, we introduce the QLT-SLC method, which outperforms existing state-of-the-art methods, enhancing both the accuracy and recall of the loop closure detection process. However, our proposed method still has some limitations, such as not considering dynamic scenes and not involving semantic objects in loop correction. Therefore, in future work, we aim to address these limitations and further enhance the robustness of the proposed method.

\bibliographystyle{IEEEtran}
\bibliography{ref}

%%%%%%%%%%%%%%%%%%%%%%%%%%%%%%%%%%%%%%%%%%%%%%%%%%%%%%%%%%%%%%%%%%%%%%%%%%%%%%%%

%%%%%%%%%%%%%%%%%%%%%%%%%%%%%%%%%%%%%%%%%%%%%%%%%%%%%%%%%%%%%%%%%%%%%%%%%%%%%%%%

\end{document}